\documentclass[letterpaper, 10 pt, journal, twoside]{IEEEtran}

\usepackage{cite}
\usepackage[pdftex]{graphicx}
\usepackage{amsmath}
\usepackage{amsthm}
\usepackage{amssymb}
\usepackage{wasysym}
\usepackage{bm}
\usepackage[]{units}
\usepackage{siunitx}
\usepackage[ruled]{algorithm2e} 
\usepackage{array}
\usepackage{url}
\usepackage{color}
\usepackage{arydshln}
\usepackage{mathtools}
\usepackage{makecell}
\usepackage{stfloats}
\usepackage{multirow}
\DeclareMathAlphabet{\mathbfcal}{OMS}{cmsy}{b}{n}

\newtheorem{definition}{Definition}
\usepackage[table,xcdraw]{xcolor}

\begin{document}
\title{Data-Efficient and Safe Learning for Humanoid Locomotion Aided by a Dynamic Balancing Model}
\author{Junhyeok Ahn$^{1}$, Jaemin Lee$^{1}$, and Luis Sentis$^{2}$
\thanks{Manuscript received: December, 19, 2019; Revised March, 9, 2020; Accepted, April 5, 2020.}
\thanks{This paper was recommended for publication by Editor Abderrahmane Kheddar upon evaluation of the Associate Editor and Reviewers' comments.
This work was supported by the Office of Naval Research, ONR Grant
\#N000141512507 and the National Science Foundation, NSF Grant \#1724360.} 
\thanks{$^{1}$J. Ahn and J. Lee are with the Department of Mechanical Engineering, the University of Texas at Austin, Austin, TX, 78712, USA {\tt\small \{junhyeokahn91, jmlee87\}@utexas.edu}}%
\thanks{$^{2}$L. Sentis is with the Department of Aerospace Engineering and Engineering Mechanics, the University of Texas at Austin, Austin, TX, 78712, USA {\tt\small lsentis@austin.utexas.edu}.}
\thanks{Digital Object Identifier (DOI): see top of this page.}
}

\markboth{IEEE Robotics and Automation Letters. Preprint Version. Accepted April, 2020}
{Ahn \MakeLowercase{\textit{et al.}}: Data-Efficient and Safe Learning for Humanoid Locomotion} 

\maketitle

\begin{abstract}
In this letter, we formulate a novel Markov Decision Process (MDP) for safe and data-efficient learning for humanoid locomotion aided by a dynamic balancing model. In our previous studies of biped locomotion, we relied on a low-dimensional robot model, commonly used in high-level Walking Pattern Generators (WPGs). However, a low-level feedback controller cannot precisely track desired footstep locations due to the discrepancies between the full order model and the simplified model. In this study, we propose mitigating this problem by complementing a WPG with reinforcement learning. More specifically, we propose a structured footstep control method consisting of a WPG, a neural network, and a safety controller. The WPG provides an analytical method that promotes efficient learning while the neural network maximizes long-term rewards, and the safety controller encourages safe exploration based on step capturability and the use of control-barrier functions. Our contributions include the following (1) a structured learning control method for locomotion, (2) a data-efficient and safe learning process to improve walking using a physics-based model, and (3) the scalability of the procedure to various types of humanoid robots and walking.
\end{abstract}


\section{Introduction and Related Work}
\label{sec:introduction}

\IEEEPARstart{H}{umanoid} robots are advantageous for mobility in tight spaces. However, fast bipedal locomotion requires precision control of the contact transition process. Many studies have successfully addressed agile and versatile legged locomotion. Analytic approaches have employed differential dynamics of robots to synthesize locomotion controllers. Data-driven approaches have leveraged the representational power of neural networks and designed locomotion policies in an end-to-end manner. Our work combines the advantages of these approaches to achieve locomotion behaviors both safely and efficiently.

Analytic approaches decouple the problem into two sub-problems: (1) reducing the complexity of full-body dynamics via simplified models, such as the inverted pendulum \cite{Kuindersma2016,Rezazadeh:vk,caron2019icra,1241826} or the centroidal model \cite{8558661,7759420,Orin2013}, to generate high-level walking patterns, and then (2) computing feedback joint commands at every control loop so that the robot tracks the behavior of the simplified models. In our recent studies \cite{dh_ijrr_2019,JunhyeokDRACO}, we achieved unsupported passive ankle dynamic locomotion via two computational elements: (1) a high-level footstep planner, called the Time-to-Velocity-Reversal (TVR) planner, based on the Linear Inverted Pendulum Model (LIPM) and (2) a low-level Whole Body Controller (WBC) that tracks the desired trajectories. Although abstractions based on simplified models enable Walking Pattern Generators (WPGs) to provide computational efficiency and tools for stability analysis, they have a limited ability to incorporate complicated physical effects, such as angular momentum and limb dynamics. As a result, using WPGs cause significant footstep tracking errors, requiring arduous parameter tuning \cite{chen2019optimal}. In this letter, we propose and train a policy that compensates for the limited representation accuracy of WPGs and generates practical walking patterns by incorporating simple physical models.

\begin{figure}[t]
    \centering
    \includegraphics[width=1.0\linewidth]{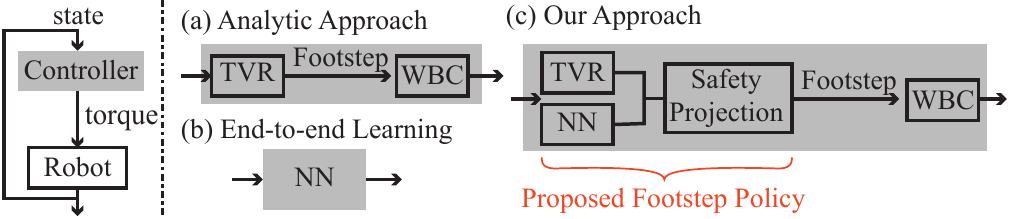}
    \caption{The left figure illustrates the control structure of a robot, where a controller takes the robot's states and computes joint torques in an end-to-end manner. (a) Analytic approaches compose the controller with a WPG (e.g., a TVR Planner) and a feedback controller (e.g., WBC), whereas (b) end-to-end learning methods train a neural network to compute the joint torques. (c) Our controller includes the footstep policy learning algorithm and WBC, where the footstep policy has three components.}
    \label{fig:intro}
\end{figure}

On the other hand, data-driven approaches have demonstrated the possibility of robust and agile locomotion control through Reinforcement Learning (RL). Model-free RL learns a walking policy via explicit trial and error without using knowledge of the dynamics of the robots. In \cite{heess2017emergence,deeploco}, locomotion policies were trained for various environments and achieved robust locomotion behaviors. In contrast, model-based RL learns a model of a robot through interactions with the environment and leverages the constructed model for planning. The approach in \cite{gps} iteratively fitted a local model for a planar walker and performed trajectory optimization, which demonstrated the ability to learn a walking policy efficiently. However, most data-driven approaches for locomotion do not consider the underlying physics of the robot nor prior knowledge and instead train policies from sensor data to joint commands in an end-to-end manner. Therefore, they require substantial training data and often result in unnatural jerky motions, which make the methods challenging to deploy in real hardware. In contrast to these end-to-end methods, our framework learns a policy from sensor data to footstep locations (instead of joint torques) and utilizes a whole-body controller to track the desired trajectories. In the footstep decision making, we rely on a LIPM and a TVR planner to encourage safe and efficient exploration in policy training.

There have been few works that incorporate physical insight and stable feedback control to learn biped locomotion. In \cite{castillo2019hybrid}, an RL agent learns a set of trajectory parameters instead of joint commands, followed by a feedback controller to stabilize the robot along the resulting trajectory. However, the policy is trained in a model-free manner and can yield infeasible trajectories that make the robot explore unsafe state-space regions. In contrast, our work proposes a structured policy with a safety mechanism as well as a TVR planner and a neural network to foster safe and efficient policy search.

Previous works have explored the idea of learning residual actions combined with analytical models. In \cite{tossing_bot}, a ball-throwing action is adjusted by a trained neural network to mitigate model discrepancies for a ball-tossing robot. In \cite{iscen2019policies}, swing foot trajectories generated by a feedback controller are modulated to improve performance and data efficiency for a quadruped. Our proposed algorithm can be seen as an extension of these ideas to bipedal robots that also includes safety considerations. Compared to robot manipulators or quadrupeds, biped robots fall more often and benefit from the use of physics-based models to guide the learning process.

In this paper, we devise a Markov Decision Process (MDP) that combines analytic models and data-driven approaches to achieve agile and robust locomotion. In contrast to end-to-end learning such as in \cite{schulman2017proximal, brockman2016openai, heess2017emergence, gps}, whose learning techniques take joint information and map them to joint torque commands, our method learns a policy to make high-level decisions in terms of desired footstep locations. It then uses a feedback whole-body controller to generate locomotion behaviors and the desired reward signals. Our structured footstep control methods includes a TVR planner, a neural network, and a safety controller. The TVR planner provides feasible sub-optimal guidance, the neural network maximizes the long-term reward, and the safety controller encourages safe exploration during the learning process. This safety controller learns the residual dynamics of the LIPM and projects the action onto safe regions considering a walking capturability metric. The overall structures of our method and those of related works are shown and compared in Fig.~\ref{fig:intro}.

The proposed MDP formulation has the following advantages: (1) it bridges the gap between analytic and data-driven approaches, which mitigates the limited effect of using simple models; (2) it allows data efficiency and safe learning; and (3) it can be used for different types of locomotion and in different types of robots.

The remainder of this paper is organized as follows. Section~\ref{sec:preliminaries} describes an analytic approach for biped locomotion and RL with safety guarantees. Section~\ref{sec:MDP_Formulation} proposes an MDP formulation for humanoid locomotion tasks. /bt{Section~\ref{sec:policy_design} shows the design of a footstep policy that allows safe exploration and data-efficient training.} Section~\ref{sec:simulation_results} evaluates the effectiveness and generalization of the proposed framework in simulation, and Section~\ref{sec:concluding_remarks} concludes the paper.

\section{Preliminaries}
\label{sec:preliminaries}



\subsection{An Analytic Approach to Locomotion}
\label{sec:wblc}

\begin{figure}
    \centering
    \includegraphics[width=1.0\linewidth]{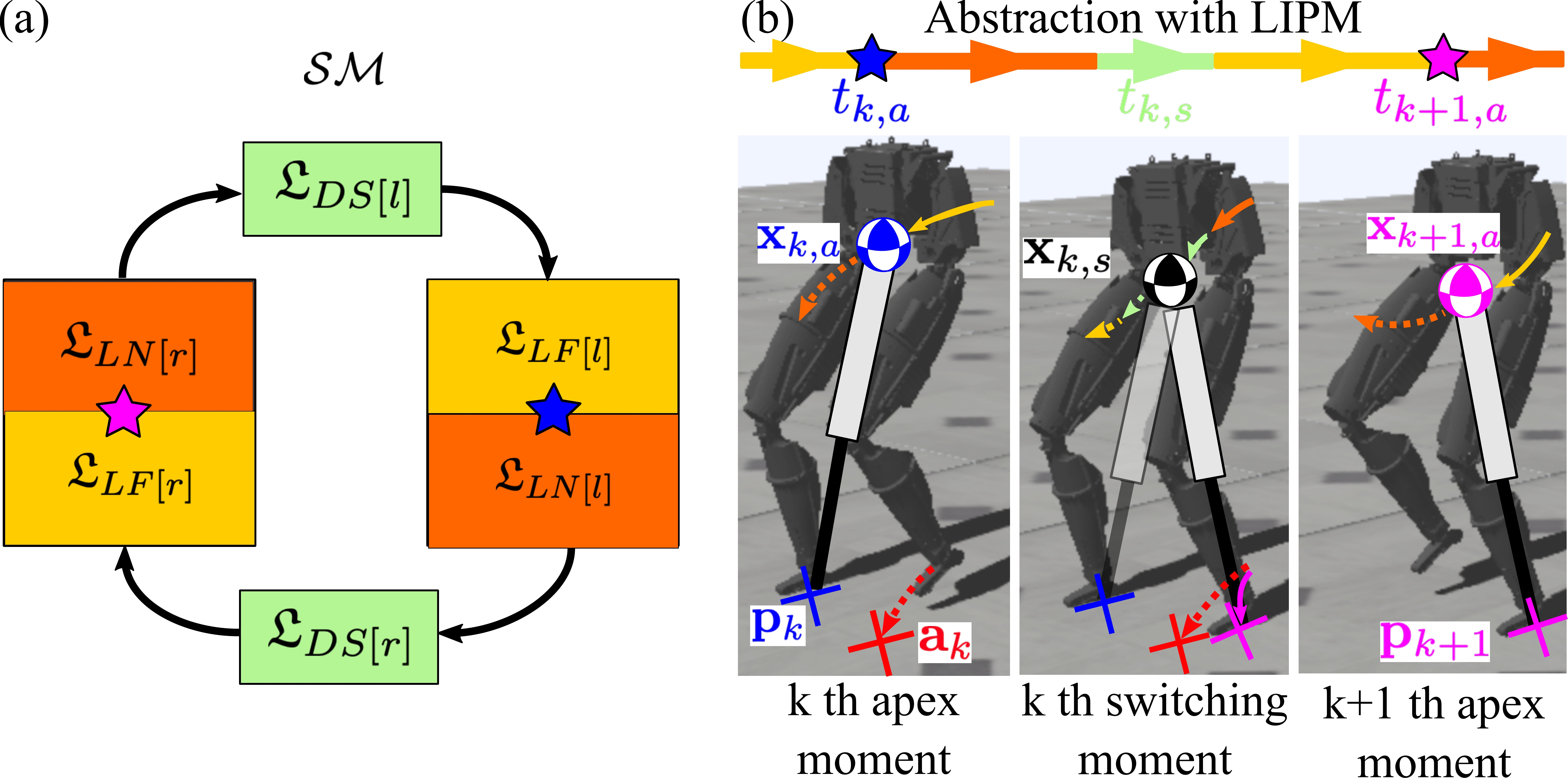}
    \caption{(a) shows the $\mathcal{SM}$ for locomotion behaviors. The blue and pink stars represent the $k$th and $k+1$th \textit{Apex Moment}s. (b) shows the walking motion with the $\mathcal{SM}$ and its abstraction using the LIPM.}
    \label{fig:preliminary}
\end{figure}

We define a \textit{Locomotion State} and a state machine with simple structures to represent general locomotion behaviors.
\begin{definition}
(\textbf{Locomotion State}) A locomotion state is defined as a tuple, $\mathfrak{L} \coloneqq ( \mathcal{L}, T_{\mathcal{L}})$.
\begin{itemize}
    \item $\mathcal{L}$ represents a semantic expression of locomotion behaviors: $\mathcal{L} \in \{\mathcal{L}_{\textrm{DS[r/l]}}, \mathcal{L}_{\textrm{LF[r]}}, \mathcal{L}_{\textrm{LF[l]}}, \mathcal{L}_{\textrm{LN[r]}}, \mathcal{L}_{\textrm{LN[l]}}\}$.
    \item The subscripts $(.)_{\textrm{DS[r/l]}}$, $(.)_{\textrm{LF[r/l]}}$, and $(.)_{\textrm{LN[r/l]}}$ describe locomotion states for double support, lifting the right/left leg, and landing the right/left leg, respectively.
    \item $T_{\mathcal{L}}$ is a time duration for $\mathcal{L}$ and can be chosen based on the desired stepping frequency.
\end{itemize}
\end{definition}
\begin{definition}
(\textbf{State Machine}) We define a state machine as a sequence of Locomotion States:
\begin{equation*}
    \mathcal{SM}\coloneqq \{ \mathfrak{L}_{\textrm{DS[r/l]}}, \mathfrak{L}_{\textrm{LF[r/l]}}, \mathfrak{L}_{\textrm{LN[r/l]}}\}
\end{equation*}
\end{definition}
\begin{itemize}
    \item The list above is sequential in the order shown.
    \item The \textit{Locomotion State} $\mathfrak{L}_{LN[r/l]}$ terminates when a contact is detected between the swing foot and the ground.
\end{itemize}

\begin{definition}
    (\textbf{Apex Moment} and \textbf{Switching Moment}) Given the $\mathcal{SM}$ defined above, an Apex Moment defines the switch between $\mathfrak{L}_{LF[r/l]}$ and $\mathfrak{L}_{\textrm{LN[r/l]}}$, and we label it as $t_a$. A Switching Moment defines the middle of $\mathfrak{L}_{\textrm{DS[r/l]}}$, and we label it as $t_{s}$.
\end{definition}

Let us consider the LIPM for our simplified model. We define the \textit{LIPM state} as the position and velocity of the Center of Mass (CoM) of the robot on a constant height surface with an expression, $\mathbf{x}=[x, y, \dot{x}, \dot{y}]^\top \in \mathbb{R}^4$. The \textit{LIPM stance} is defined as the location of the pivot and represented by $\mathbf{p}=[p_x, p_y]^\top \in \mathbb{R}^{2}$. We define the \textit{LIPM input} as the desired location of the next stance with an expression $\mathbf{a}=[a_x, a_y]^\top \in \mathbb{R}^2$.  We use the subscript $k$ to represent properties in the $k$th step, for example, $\mathbf{x}_k=[x_k, y_k, \dot{x}_{k}, \dot{y}_{k}]^\top$, $\mathbf{p}_{k}=[p_{k,x}, p_{k,y}]^\top$, and $\mathbf{a}_{k}=[a_{k,x},a_{k,y}]^\top$. We further use the subscripts $k,a$, and $k,s$ to denote the properties of the robot at the \textit{Apex Moment} and the \textit{Switching Moment} at the $k$th step. For example, $\mathbf{x}_{k,*} = [x_{k,*}, y_{k,*}, \dot{x}_{k,*}, \dot{y}_{k,*}]^\top = \mathbf{x}_k(t_{k,*})$, where $* \in \{a, s\}$ represents the \textit{LIPM state} evaluated at the \textit{Apex Moment} and \textit{Switching Moment} at the $k$th step. Because the \textit{LIPM stance} and \textit{LIPM input} are invariant during the step, $\mathbf{p}_{k,a}$ and $\mathbf{a}_{k,a}$ are interchangeable with $\mathbf{p}_{k}$ and $\mathbf{a}_{k}$. We also use these subscripts to describe the properties of a robot. For instance, $\bm{\phi}^{\rm{bs}}_{k,a} \in SO(3)$ and $\mathbf{w}^{\rm{bs}}_{k,a} \in \mathbb{R}^3$ represent the orientation and angular velocity of a base link, respectively, and $\bm{\phi}^{\rm{pv}}_{k,a} \in SO(3)$ represents the orientation of a stance foot (a pivot) with respect to the world frame at the \textit{Apex Moment} at the $k$th step. Fig.~\ref{fig:preliminary} illustrates the $\mathcal{SM}$ and the abstraction of the locomotion behavior with the LIPM.

The goal of the WPG is to generate $\mathbf{a}_{k}$ and the CoM trajectory based on $\mathbf{x}_{k,a}$ and $\mathbf{p}_{k}$ at the \textit{Apex Moment}. From the walking pattern, the low-level WBC provides the computation of sensor-based feedback control loops and torque command for the robot to track the desired location of the next stance and the CoM trajectory.  Note that the WPG designs the pattern at the \textit{Apex Moment} at each step, while the WBC computes the feedback torque command at every control loop.

\subsection{TVR Planner}
\label{sec:foot_placement_planner}

The differential equation of the LIPM is represented as follows:
\begin{equation}
    \label{eq:lipm_dyn}
    \mathbf{\dot{x}}(t) = 
    \begin{bmatrix}
        0   &   0   &   1   &   0\\
        0   &   0   &   0   &   1\\
        g/h &   0   &   0   &   0\\
        0   &   g/h &   0   &   0
    \end{bmatrix} \mathbf{x}(t) -
    \begin{bmatrix}
        0   &   0\\
        0   &   0\\
        g/h &   0\\
        0   &   g/h
    \end{bmatrix} \mathbf{p},
\end{equation}
where $g$ is the gravitational constant and $h$ is the constant height of the CoM of the point mass.

At the $k$th step, given an initial condition $\mathbf{x}_{k}(0)=\mathbf{x}_{k,0}$ and a stance position $\mathbf{p}_k$, the solution of Eq.~\eqref{eq:lipm_dyn} yields a state transition map $\Psi$, with the expression
\begin{equation}
    \label{eq:state_transition_function}
    \mathbf{x}_{k}(t) = \Psi(t\: ; \: \mathbf{x}_{k, 0}, \mathbf{p}_{k}) = f_{\Psi}(t) \mathbf{x}_{k,0} + g_{\Psi}(t)\mathbf{p}_{k},
\end{equation}
where
\begin{equation*}
\begin{split}    
    f_{\Psi}(t) &\coloneqq 
         \begin{bmatrix}
            C_1(t) & 0 & C_2(t) & 0\\
            0 & C_1(t) & 0 & C_2(t)\\
            C_3(t) & 0 & C_1(t) & 0 \\
            0 & C_{3}(t) & 0 & C_1(t)
         \end{bmatrix}, \\
    g_{\Psi}(t) & \coloneqq
         \begin{bmatrix}
            1-C_1(t) & 0 \\
            0 & 1-C_1(t) \\
            -C_3(t) & 0\\
            0 & -C_3(t)
         \end{bmatrix},
\end{split}
\end{equation*}
$C_1(t) \coloneqq \cosh(\omega t)$, $C_2(t) \coloneqq \sinh(\omega t )/\omega$, and $C_3(t) \coloneqq \omega \sinh(\omega t)$, and $\omega \coloneqq \sqrt{g/h}$, respectively.

Because the TVR planner determines the desired location of the next stance at the \textit{Apex Moment} (i.e., $t=t_{k,a}$), we set the initial condition as $\mathbf{x}_{k}(0) = \mathbf{x}_{k,a}$. With pre-specified time duration $T_{\mathcal{L}_{LN[r/l]}}$, we compute the state at the \textit{Switching Moment} as
\begin{equation}
    \mathbf{x}_{k,s} = \mathbf{x}_{k}(T_{\mathcal{L}_{LN[r/l]}}) = \Psi(T_{\mathcal{L}_{LN[r/l]}} \: ; \: \mathbf{x}_{k}(t_{k,a}), \mathbf{p}_{k}).
\end{equation}
From $\mathbf{x}_{k,s}$, the TVR planner computes $\mathbf{a}_{k}$, such that the sagittal velocity $\dot{x}$ (and lateral velocity $\dot{y}$, respectively) of the CoM is driven to zero at the predefined time intervals $T_{x'}$ (and $T_{y'}$, respectively) after the LIPM switches to the new stance. These constraints are expressed as
\begin{equation}
    \label{eq:velocity_reversal_constraint}
        0 = \big\langle \bm{\xi}_{j} \:,\: \Psi(T_{j} \: ; \: \mathbf{x}_{k,s},\mathbf{a}_{k}) \big\rangle ,\quad j \in \{ x', y' \},
\end{equation}
where $\bm{\xi}_{x'} \coloneqq [0, 0, 1 , 0]^\top$ and $\bm{\xi}_{y'}\coloneqq [0, 0, 0, 1]^\top$. From Eq.~\eqref{eq:velocity_reversal_constraint}, $\mathbf{a}_{k}$ is computed with an additional bias term $\kappa_{x}$ and $\kappa_{y}$ as
\begin{equation}
    \label{eq:tvr}
    \mathbf{a}^{\rm{TVR}}_{k} = \Phi(\mathbf{x}_{k,s}) = f_{\Phi}(T_{x'},T_{y'})\mathbf{x}_{k,s} + g_{\Phi},
\end{equation}
where
\begin{equation*}
    \begin{split}
    f_{\Phi}(T_{x'},T_{y'}) \coloneqq& \begin{bmatrix}
        1 - \kappa_{x} & 0 & C_4(T_{x'}) & 0 \\
        0 & 1 - \kappa_{y} & 0 & C_4(T_{y'})
    \end{bmatrix}, \\
    g_{\Phi} \coloneqq& \begin{bmatrix} \kappa_{x} & 0 \\ 0 & \kappa_{y}\end{bmatrix} \begin{bmatrix} x^d \\ y^d \end{bmatrix},
    \end{split}
\end{equation*}
$C_4(T) \coloneqq \frac{e^{w T} + e^{-w T}}{w(e^{w T} - e^{-w T})}$ and $[x^d,y^d]^\top \in \mathbb{R}^2$ represents a desired position for the CoM of the robot. Note that Eq.~\eqref{eq:tvr} is a simple proportional-derivative controller and that $T_{x'}, T_{y'}, \kappa_{x}$, and $\kappa_{y}$ are the gain parameters used to keep the CoM converging to the desired position. A more detailed derivation of the LIPM was described in \cite{ahn_2018}.


\subsection{Reinforcement Learning with Safe Exploration}
\label{sec:reinforcement_learning_with_safety_guarantee}

Consider an infinite-horizon discounted MDP with control-affine, deterministic dynamics defined by the tuple $(\mathcal{S},\mathcal{A},\mathcal{T}, r, \rho_0, \gamma)$, where $\mathcal{S}$ is a set of states, $\mathcal{A}$ is a set of actions, $\mathcal{T} : S \mapsto S$ is the deterministic dynamics, in our case affine in the controls, $r : \mathcal{S} \times \mathcal{A} \mapsto \mathbb{R}$ is the reward function, $\rho_0 : \mathcal{S} \mapsto \mathbb{R}$ is the distribution of the initial state, and $\gamma \in (0, 1)$ is the discount factor. The control affine dynamics are written as
\begin{equation}
    \mathbf{s}_{k+1} = f(\mathbf{s}_k) + g(\mathbf{s}_k)\mathbf{a}_k + d(\mathbf{s}_k),
\end{equation}
where $\mathbf{s}_k \in \mathcal{S} \subseteq \mathbb{R}^{n_s}$, and $\mathbf{a}_k \in \mathcal{A} \subset \mathbb{R}^{n_a}$ represent a state and input, respectively.  $f:\mathcal{S} \mapsto \mathcal{S}$, and $g : \mathcal{S} \mapsto \mathbb{R}^{n_s \times n_a}$ are the analytic underactuated and actuated dynamics, respectively, while $d : \mathcal{S} \mapsto \mathcal{S}$ is the unknown part of the system dynamics.  Moreover, let $\pi_{\bm{\theta}}(\mathbf{a} \vert \mathbf{s})$ represent a stochastic control policy parameterized by a vector $\bm{\theta}$.  $\pi_{\bm{\theta}}:\mathcal{S} \times \mathcal{A} \mapsto \mathbb{R}_{\geq 0}$ maps states to distributions over actions, and $V_{\pi_{\bm{\theta}}}(\mathbf{s})$ represents the policy's expected discounted reward with the expression
\begin{equation}
    \label{eq:return}
    V_{\pi_{\bm{\theta}}}(\mathbf{s}_k) = \mathbb{E}_{\tau \sim \pi_{\bm{\theta}}}\left[\sum_{i=0}^{\infty}\gamma^i r(\mathbf{s}_{k+i}, \mathbf{a}_{k+i})\right],
\end{equation}
where $\tau \sim \pi_{\bm{\theta}}$ is a trajectory drawn from the policy $\pi_{\bm{\theta}}$ (e.g., $\tau = [\mathbf{s}_k, \mathbf{a}_k, \cdots, \mathbf{s}_{k+n}, \mathbf{a}_{k+n}]$).

For safe exploration in the learning process under uncertain dynamics, the work in \cite{cheng2019end} employed a Gaussian Process (GP) to approximate the unknown part of the dynamics from the dataset by learning a mean estimate $\bm{\mu}_d(\mathbf{s})$ and an uncertainty $\bm{\sigma}_d^2(\mathbf{s})$ in tandem with the policy update with probability confidence intervals on the estimation,
\begin{equation}
    \label{eq:gp}
    \bm{\mu}_d(\mathbf{s}) - k_\delta \bm{\sigma}_d (\mathbf{s}) \leq d(\mathbf{s}) \leq \bm{\mu}_d(\mathbf{s}) + k_\delta \bm{\sigma}_d(\mathbf{s}),
\end{equation}
where $k_\delta$ is a design parameter indicating a confidence. Then, the control input is computed to keep the following state within a given invariant set $\mathcal{C} = \{ \mathbf{s} \in \mathcal{S} \: \vert \: h(\mathbf{s}) \geq 0 \}$ by computing
\begin{equation}
    \label{eq:cbf_opt}
    \sup_{\mathbf{a_k} \in \mathcal{A}} \big[ h\big( f(\mathbf{s}_k) + g(\mathbf{s}_k) \mathbf{a}_k + d(\mathbf{s}_k) \big) + (\eta - 1)h(\mathbf{s}_k) \big] \geq 0,
\end{equation}
where $\eta \in [0,1]$.

\section{MDP Formulation}
\label{sec:MDP_Formulation}

We define a set of states $\mathcal{S}$ and a set of actions $\mathcal{A}$ associated with the \textit{Apex Moment} at each step:
\begin{equation*}
\begin{split}
    \mathcal{S} &\coloneqq \big\{(\mathbf{x}_{k,a}, \mathbf{p}_{k,a}, \bm{\phi}^{\rm{bs}}_{k,a}, \mathbf{w}^{\rm{bs}}_{k,a}, \bm{\phi}^{\rm{pv}}_{k,a}) \: \vert \: \forall k \in [1, m]_{\mathbb{N}} \big\}, \\
    \mathcal{A} &\coloneqq \big\{\mathbf{a}_{k,a} \: \vert \: \forall k \in [1, m]_{\mathbb{N}} \big\},
\end{split}
\end{equation*}
where $m$ can be set as $+\infty$ when considering the infinite steps of the locomotion. Recall from the nomenclatures in Section~\ref{sec:wblc} that $\mathbf{x}_{k,a},~\mathbf{p}_{k,a}$ and $\mathbf{a}_{k,a}$ are the expressions of the \textit{LIPM state}, \textit{LIPM stance}, and \textit{LIPM input} evaluated at the \textit{Apex Moment}.  Note that $\mathbf{p}_{k,a}$ and $\mathbf{a}_{k,a}$ are interchangeable with $\mathbf{p}_{k}$ and $\mathbf{a}_{k}$. Moreover, $\bm{\phi}^{\rm{bs}}_{k,a}$ and $\mathbf{w}^{\rm{bs}}_{k,a}$ represent the orientation and angular velocity of a base link and $\bm{\phi}^{\rm{pv}}_{k,a}$ expresses an orientation of the stance foot at the \textit{Apex Moment}. We divide the state into two parts as
\begin{equation}
    \label{eq:state}
    \begin{split}
        \mathbf{s}_{k+1} &= \left[
    \begin{array}{c|c}
        \mathbf{s}_{k+1}^{u} & \mathbf{s}_{k+1}^{l}
    \end{array}\right]^{\top} \\
    &=
    \left[
    \begin{array}{c c | c c c}
        \mathbf{x}_{k+1,a} & \mathbf{p}_{k+1} &  \bm{\phi}^{\rm{bs}}_{k+1,a} & \mathbf{w}^{\rm{bs}}_{k+1,a} & \bm{\phi}^{\rm{pv}}_{k+1,a}
    \end{array}\right]^{\top}
    \end{split}
\end{equation}
and define a transition function for the upper part of the state based on Eq.~\eqref{eq:state_transition_function} as
\begin{equation} \label{eq:transition_map}
    \begin{split}
        \mathbf{s}_{k+1}^{u} =& f(\mathbf{x}_{k,a},\mathbf{p}_{k}) + g \mathbf{a}_k + d(\mathbf{x}_{k,a},\mathbf{p}_{k}),\\
        f(\mathbf{x}_{k,a},\mathbf{p}_{k}) \coloneqq& \left[ \begin{array}{c}
            f_\Psi(T_{LF}) \Psi(T_{LN}; \mathbf{x}_{k, a}, \mathbf{p}_k)   \\
            \mathbf{0}_{2 \times 1}  
        \end{array} \right], \\
        g \coloneqq& \left[ \begin{array}{c}
             g_\Psi(T_{LF})  \\ \mathbf{I}_{2 \times 2}
        \end{array} \right].
    \end{split}
\end{equation}
$d(\mathbf{x}_{k,a},\mathbf{p}_{k})$ represents the unknown part of the dynamics fitted via Eq.~\eqref{eq:gp}\footnote{We use a squared exponential kernel for GP prior to implementation.}. The uncertainties are attributed to the discrepancies between the simplified model and the actual robot. Note that the dynamics of the lower part of the states, $\mathbf{s}^{l}_{k+1}$, cannot be expressed in closed form. Therefore, we optimize our policy in a model-free sense, but utilize the LIPM to provide safe exploration and data efficiency in the learning process.

To train a policy for a locomotion behavior, we adapt a reward function from \cite{brockman2016openai}, widely-used for locomotion tasks:
\begin{equation}
    \label{eq:rew}
    r(\mathbf{s}_k, \mathbf{a}_k) = r_a + r_b(\mathbf{s_k}) + r_t(\mathbf{s_k}) + r_s(\mathbf{s_k}) + r_c(\mathbf{a_k}).
\end{equation}
Given $\mathbf{w}^{\rm{bs}}_{k,a} = [w_{k,x},w_{k,y},w_{k,z}]^\top$, the Euler ZYX representation $[\bm{\phi}^{\rm{bs}}_{k,x}, \bm{\phi}^{\rm{bs}}_{k,y}, \bm{\phi}^{\rm{bs}}_{k,z}]^\top$ of $\bm{\phi}^{\rm{bs}}_k$ and $[\bm{\phi}^{\rm{pv}}_{k,x}, \bm{\phi}^{\rm{pv}}_{k,y}, \bm{\phi}^{\rm{pv}}_{k,z}]^\top$ of $\bm{\phi}^{\rm{pv}}$, $r_a$ is an alive bonus, $r_b(\mathbf{s}_k) \coloneqq - w_b \Vert (\bm{\phi}^{\rm{bs}}_{k,x}, \bm{\phi}^{\rm{bs}}_{k,y}) \Vert^2$ penalizes the roll and pitch variation to keep the body upright, $r_t(\mathbf{s}_k) \coloneqq - w_t \Vert (x^d_{k,a}, y^d_{k,a}, \bm{\phi}^{\rm{bs},d}_{k,z}, \bm{\phi}^{\rm{pv},d}_{k,z}) - (x_k, y_k, \bm{\phi}^{\rm{bs}}_{k,z}, \bm{\phi}^{\rm{pv}}_{k,z}) \Vert^2$ penalizes divergence from the desired CoM positions and the heading of the robot, $r_s(\mathbf{s}_t) \coloneqq - w_s \Vert (\dot{x}^d_{k,a}, \dot{y}^d_{k,a}, w^d_{k,z}) - (\dot{x}_{k,a}, \dot{y}_{k,a}, w_{k,z}) \Vert^2$ is for steering the robot with a desired velocity, and $r_c(\mathbf{a}_k) \coloneqq -w_c \Vert \mathbf{a}_k \Vert^2$ penalizes excessive control input.

\section{Policy Representation and Learning}
\label{sec:policy_design}

Our goal is to learn an optimal policy for desired foot locations. We use the Proximal Policy Optimization (PPO) \cite{schulman2017proximal} to learn the policy iteratively. PPO defines an advantage function  $A_{\pi_{\bm{\theta}}}(\mathbf{s}_k, \mathbf{a}_k) \coloneqq Q_{\pi_{\bm{\theta}}}(\mathbf{s}_k, \mathbf{a}_k) - V_{\pi_{\bm{\theta}}}(\mathbf{s}_k)$, where $Q_{\pi_{\bm{\theta}}}(\mathbf{s}_k, \mathbf{a}_k)$ is the state-action value function that evaluates the return of taking action $\mathbf{a}_k$ at state $\mathbf{s}_k$ and following the policy $\pi$ thereafter. By maximizing a modified objective function
\begin{equation*}
    L_{\textrm{PPO}}(\bm{\theta}) = \mathbb{E}_{\tau \sim \pi_{\bm{\theta}}}\left[ \min \big( r_k A_k, clip(r_k, \: 1-\epsilon, \: 1+\epsilon)A_k \big) \right],
\end{equation*}
where $r_k \coloneqq \frac{\pi_{\bm{\theta}}(\mathbf{a}_k \vert \mathbf{s}_k)}{\pi_{\bm{\theta}_{\rm{old}}}(\mathbf{a}_k \vert \mathbf{s}_k)}$ is the importance resampling term that allows us to use the dataset under the old policy $\pi_{\bm{\theta}_{\rm{old}}}$ to estimate for the current policy $\pi_{\bm{\theta}}$. $A_k$ is a short notation for $A_{\pi_{\bm{\theta}}}(\mathbf{s}_k, \mathbf{a}_k)$. The $\min$ and $clip$ operator ensures that the policy $\pi_{\bm{\theta}}$ does not change excessively from the old policy $\pi_{\bm{\theta}_{\rm{old}}}$.

\subsection{Safe Set Approximation}

The work in \cite{captur_point} introduced an instantaneous capture point that enables the LIPM to come to a stop if it places and maintains its stance there instantaneously. Here, we consider $i$-step capture regions for the LIPM at the \textit{Apex Moment}:
\begin{equation} \label{eq:true_one_step_region}
        \left \Vert \begin{bmatrix} 1 & 0 & 1/\omega & 0 & -1 & 0 \\ 0 & 1 & 0 & 1/\omega & 0 & -1\end{bmatrix}\ \begin{bmatrix} \mathbf{x}_{k,a} \\ \mathbf{p}_{k} \end{bmatrix} \right \Vert \leq \mathcal{CP}_i,
\end{equation}
where 
\begin{equation*}
    \begin{split}
        \mathcal{CP}_{1} &\coloneqq l_{\rm{max}} e^{-T_{\mathcal{L}_{LN[r/l]}}}, \\
        \mathcal{CP}_{2} &\coloneqq l_{\rm{max}} e^{-T_{\mathcal{L}_{LN[r/l]}}} (1+e^{-T_{\mathcal{L}_{LN[r/l]}}}),
    \end{split}
\end{equation*}
$\omega = \sqrt{g/h}$, and $l_{\rm{max}}$ is the maximum step length that the LIPM can reach. Both $\omega$ and $l_{\rm{max}}$ are achieved from the kinematics of a robot. $T_{\mathcal{L}_{LN[r/l]}}$ is a predefined temporal parameter that represents the time period until the robot lands its swing foot. We conservatively approximate the ellipsoid of Eq.~\eqref{eq:true_one_step_region} with a polytope and define a safe set of states as
\begin{equation}
\begin{split}
    \label{eq:safe_set_lipm}
    \mathcal{C} = \big\{(\mathbf{x}_{k,a},\mathbf{p}_{k})\:\vert\: h(\mathbf{x}_{k,a},\mathbf{p}_{k}) \geq \mathbf{0}_{4 \times 1}, \forall k \in [1, m]_{\mathbb{N}}\big\},
\end{split}
\end{equation}
where
\begin{equation} \label{eq:h}
    \begin{split}
        &h(\mathbf{x}_{k,a}, \mathbf{p}_{k}) \coloneqq \mathbf{A}_{\mathcal{C}} \begin{bmatrix} \mathbf{x}_{k,a} \\ \mathbf{p}_{k} \end{bmatrix} + \mathbf{b}_{\mathcal{C}}, \\
            &\mathbf{A}_{\mathcal{C}} \coloneqq \frac{1}{\mathcal{CP}_{i}}
        \begin{bmatrix}
            -1& -1& -1/\omega & -1/\omega & 1 & 1 \\
            -1& 1& -1/\omega & 1/\omega & 1 & -1\\
            1& -1& 1/\omega & -1/\omega & -1 & 1 \\
            1& 1& 1/\omega & 1/\omega & -1 & -1
        \end{bmatrix}, \\
        &\mathbf{b}_{\mathcal{C}} \coloneqq \mathbf{1}_{4 \times 1}.
    \end{split}
\end{equation}
The safe set of states in Eq.~\eqref{eq:safe_set_lipm} represents the set of the \textit{LIPM state} and \textit{LIPM stance} pairs that could be stabilized without falling by taking $i$-step. In other words, if an \textit{LIPM state} and \textit{LIPM stance} pair is inside the safe set at the $k$th step, there is always a location for the next stance $\mathbf{a}_{k}$ (and the following stance $\mathbf{a}_{k+1}$ in the case of two-step capture region) that stabilizes the LIPM. The projection onto the $x$ and $\dot{x}$ plane of capture regions is represented in Fig.~\ref{fig:action_design}(b).

\subsection{Safety Guaranteeing Policy Design}
\begin{figure}[t]
    \centering
    \includegraphics[width=1.0\linewidth]{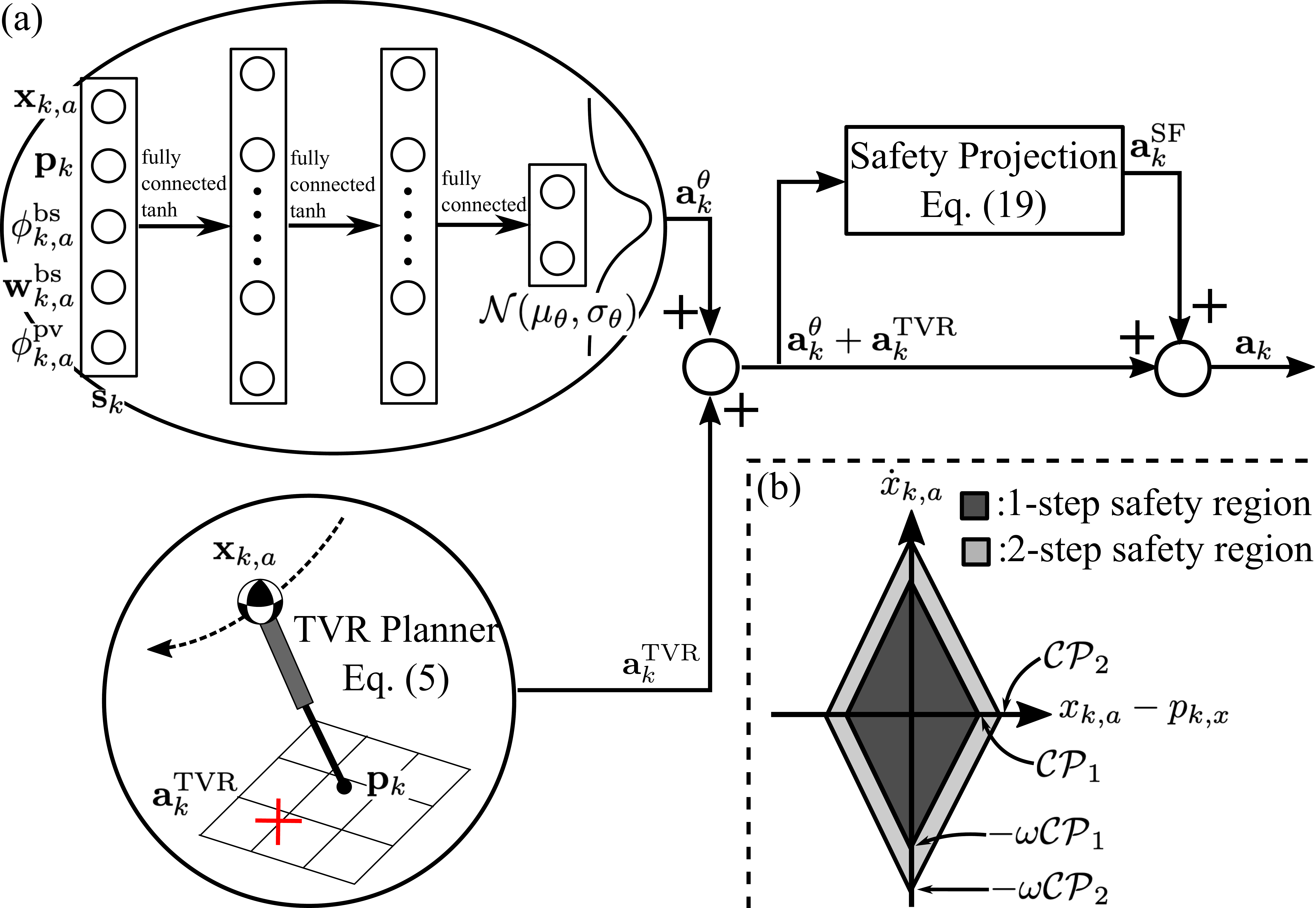}
    \caption{(a) The design of the safety-guaranteeing policy, $\mathbf{a}_k$. (b) The projection onto the $x$ and $\dot{x}$ plane of the one- and two-step capture regions of the LIPM.}
    \label{fig:action_design}
\end{figure}
For data-efficient and safe learning, we design our control input with three components:
\begin{equation}
    \label{eq:action_design}
    \mathbf{a}_k = \mathbf{a}^{\rm{TVR}}_k + \mathbf{a}^{\bm{\theta}}_k + \mathbf{a}^{\rm{SF}}_k(\mathbf{a}^{\rm{TVR}}_k + \mathbf{a}^{\bm{\theta}}_k),
\end{equation}
where $\mathbf{a}^{\rm{TVR}}_k = \Phi\big( \Psi(T_{LN}, 0 \:;\:\mathbf{x}_k, \mathbf{p}_k) \big)$ is computed by the TVR planner and $\mathbf{a}^{\bm{\theta}}_k$ is drawn from a parameterized Gaussian distribution, $\mathcal{N}(\bm{\mu}_{\bm{\theta}},\bm{\sigma}_{\bm{\theta}})$, where $\bm{\mu}_{\bm{\theta}}$ and $\bm{\sigma}_{\bm{\theta}}$ denote a mean vector and covariance matrix parameterized by $\bm{\theta}$\footnote{In the implementation, we choose two fully connected hidden layers with the $\tanh$ activation function.}, respectively. $\mathbf{a}_{k}^{\rm{SF}}:\mathcal{A} \mapsto \mathcal{A}$ is the safety projection that takes the sum of $\mathbf{a}_{k}^{\rm{TVR}}$ and $\mathbf{a}_{k}^{\bm{\theta}}$ and computes a compensation to make the action safe. Given arbitrary $\mathbf{a}^{\rm{TVR}}_k$ and $\mathbf{a}^{\bm{\theta}}_k$, the safety-guaranteeing controller $\mathbf{a}^{\rm{SF}}_k$ ensures the following \textit{LIPM state} and \textit{LIPM stance} pair ($\mathbf{x}_{k+1,a},\mathbf{p}_{k+1}$) steered by the final control input ($\mathbf{a}_k$) stays inside the safe set $\mathcal{C}$. In our problem, Eq.~\eqref{eq:cbf_opt} is modified as
\begin{equation} \label{eq:cbf_opt2}
    \sup_{\mathbf{a_k^{\rm{SF}}}} \big[ h( \mathbf{x}_{k+1,a}, \mathbf{p}_{k+1} ) + (\eta - 1)h(\mathbf{x}_{k,a}, \mathbf{p}_{k}) \big] \geq \mathbf{0}_{4 \times 1}.
\end{equation}
Substituting Eq.~\eqref{eq:transition_map} and Eq.~\eqref{eq:h} into Eq.~\eqref{eq:cbf_opt2}, and choosing the worst case for the uncertain dynamics in Eq.~\eqref{eq:gp} yields the following inequality constraint:
\begin{equation} \label{eq:safe_constraint}
    \begin{split}
        \mathbf{A}_{\mathcal{C}}\Big(f(\mathbf{x}_{k,a}, \mathbf{p}_{k}) + g(\mathbf{a}_{k}^{\rm{TVR}}+\mathbf{a}_{k}^{\bm{\theta}}+\mathbf{a}_{k}^{\rm{SF}}) + \bm{\mu}_{d}(\mathbf{x}_{k,a}, \mathbf{p}_{k}) \\ - |k_{\delta} \bm{\sigma}_{d}(\mathbf{x}_{k,a}, \mathbf{p}_{k})|\Big) + \mathbf{b}_{\mathcal{C}} \geq (1-\eta) (\mathbf{A}_{\mathcal{C}}\begin{bmatrix}\mathbf{x}_{k,a} \\ \mathbf{p}_{k}\end{bmatrix} + \mathbf{b}_{\mathcal{C}}).
    \end{split}
\end{equation}

Considering the safety constraint in Eq.~\eqref{eq:safe_constraint} and input boundaries, the optimization problem is summarized in the following Quadratic Programming (QP) and efficiently solved for the safety compensation as
\begin{equation}
\begin{split}
\label{eq:qp}
    \min_{\mathbf{a}^{\rm{SF}}_k, \epsilon}& \quad \Vert \mathbf{a}^{\rm{SF}}_k \Vert + K_\epsilon \epsilon \\
    \textrm{s.t.}& \quad
 \left[
    \begin{array}{cc}
        \mathbf{A}_{\rm{qp}}^{(11)} &\mathbf{A}_{\rm{qp}}^{(12)} \\
        \mathbf{A}_{\rm{qp}}^{(21)} &\mathbf{A}_{\rm{qp}}^{(22)} \\
        \mathbf{A}_{\rm{qp}}^{(31)} &\mathbf{A}_{\rm{qp}}^{(32)}
    \end{array}
    \right]
    \begin{bmatrix}
        \mathbf{a}^{\rm{SF}}_k \\ \epsilon 
    \end{bmatrix} 
        \leq
        \left[ \begin{array}{c}
        \mathbf{b}_{\rm{qp}}^{(1)} \\ \mathbf{b}_{\rm{qp}}^{(2)} \\ \mathbf{b}_{\rm{qp}}^{(3)}
        \end{array}\right],
\end{split}
\end{equation}
\begin{algorithm}[t]
\label{code:efficient_safe_learning}
 \KwData{$M$ episodes, $K$ data samples}
 \KwResult{$\pi_{\bm{\theta}}$}
 Initialize $\pi_{\bm{\theta}}$, $\mathbf{s}_{0} \sim \rho_0$, data array $D$ \;
 \For{$m=1:M$}{
    \For{$k=1:K$}{
        $\mathbf{a}^{\bm{\theta}}_k \sim \pi_{\bm{\theta}}$,
        ~ $\mathbf{a}^{\rm{TVR}}_{k} \gets$ Eq.~\eqref{eq:tvr} \;
        $\mathbf{a}^{\rm{SF}}_{k} \gets$ Eq.~\eqref{eq:qp} \;
        $\mathbf{a}_k \gets$ Eq.~\eqref{eq:action_design} \;
        $r_k \gets$ Eq.~\eqref{eq:rew} \;
        $\mathbf{s}_{k+1,a} \gets $ WBC stabilizes the robot and brings it to the next \textit{Apex Moment}\;
        store $(\mathbf{s}_{k},\mathbf{a}_k,\mathbf{s}_{k+1},r_{k})$ in $D$ \;
    }
    $\pi_{\theta} \gets$ Optimize $L_{\rm{PPO}}$ with $D$ w.r.t $\bm{\theta}$ \;
    Update GP model with $D$ \;
    clear $D$ \;
 }
 \caption{Policy Learning Process}
\end{algorithm}
where $\epsilon$ is a slack variable in the safety constraint, and $K_\epsilon$ is a large constant to penalize safety violation. 
Here,
\begin{alignat*}{3}
    &\mathbf{A}_{\rm{qp}}^{(11)} = -\mathbf{A}_{\mathcal{C}}g,
    & \quad \mathbf{A}_{\rm{qp}}^{(12)} = - \mathbf{1}_{4\times 1},
    & \quad \mathbf{A}_{\rm{qp}}^{(21)} = \mathbf{I}_{2 \times 2}, \\
    & \mathbf{A}_{\rm{qp}}^{(22)} = \mathbf{0}_{2\times 1},
    & \quad \mathbf{A}_{\rm{qp}}^{(31)} = -\mathbf{I}_{2 \times 2},
    & \quad \mathbf{A}_{\rm{qp}}^{(32)} = \mathbf{0}_{2\times 1},
\end{alignat*}
and
\begin{equation*}
    \begin{split}
        \mathbf{b}_{\rm{qp}}^{(1)} = &\mathbf{A}_{\mathcal{C}}\big(f(\mathbf{x}_{k,a},\mathbf{p}_k) + \bm{\mu}_d(\mathbf{x}_{k,a},\mathbf{p}_k) + g(\mathbf{a}^{\rm{TVR}}_k + \mathbf{a}^{\bm{\theta}}_k\big) \\ &- (1-\eta)\mathbf{A}_{\mathcal{C}}\begin{bmatrix} \mathbf{x}_{k,a} \\ \mathbf{p}_k\end{bmatrix} - k_{\delta}\vert \mathbf{A}_{\mathcal{C}} \bm{\sigma}_{d}(\mathbf{x}_{k,a},\mathbf{p}_k)\vert + \eta \mathbf{b}_{\mathcal{C}}, \\
        \mathbf{b}_{\rm{qp}}^{(2)} = &-\big(\mathbf{a}^{\rm{TVR}}_k + \mathbf{a}_k^{\bm{\theta}}\big) + \mathbf{a}_{\rm{max}} \\
        \mathbf{b}_{\rm{qp}}^{(3)} = &\big(\mathbf{a}^{\rm{TVR}}_k + \mathbf{a}_k^{\bm{\theta}}\big) - \mathbf{a}_{\rm{min}}.
    \end{split}
\end{equation*}

The first segment of the inequality represents a constraint for the safety, and the last two are for the input constraints. The design of the safety-guaranteeing policy is illustrated in Fig.~\ref{fig:action_design}(a). Based on the MDP formulation and the policy design, the overall algorithm for efficient and safe learning for locomotion behaviors is summarized in Alg.~\ref{code:efficient_safe_learning}.

\subsection{Further Details}
\label{sec:further_details}

It is worth taking a look at each of the components in the final action described by Eq.~\eqref{eq:action_design}. $\mathbf{a}^{\rm{TVR}}_{k} + \mathbf{a}^{\bm{\theta}}_{k}$ provides a ``feedforward exploration'' in the state space, where the stochastic action explores the TVR planner and optimizes the long-term reward. $\mathbf{a}_k^{\rm{SF}}$ projects $\mathbf{a}^{\rm{TVR}}_{k} + \mathbf{a}^{\bm{\theta}}_{k}$ onto the safe set of policies and furnishes ``safety compensation''.

\begin{figure}[t]
    \centering
    \includegraphics[width=0.9\linewidth]{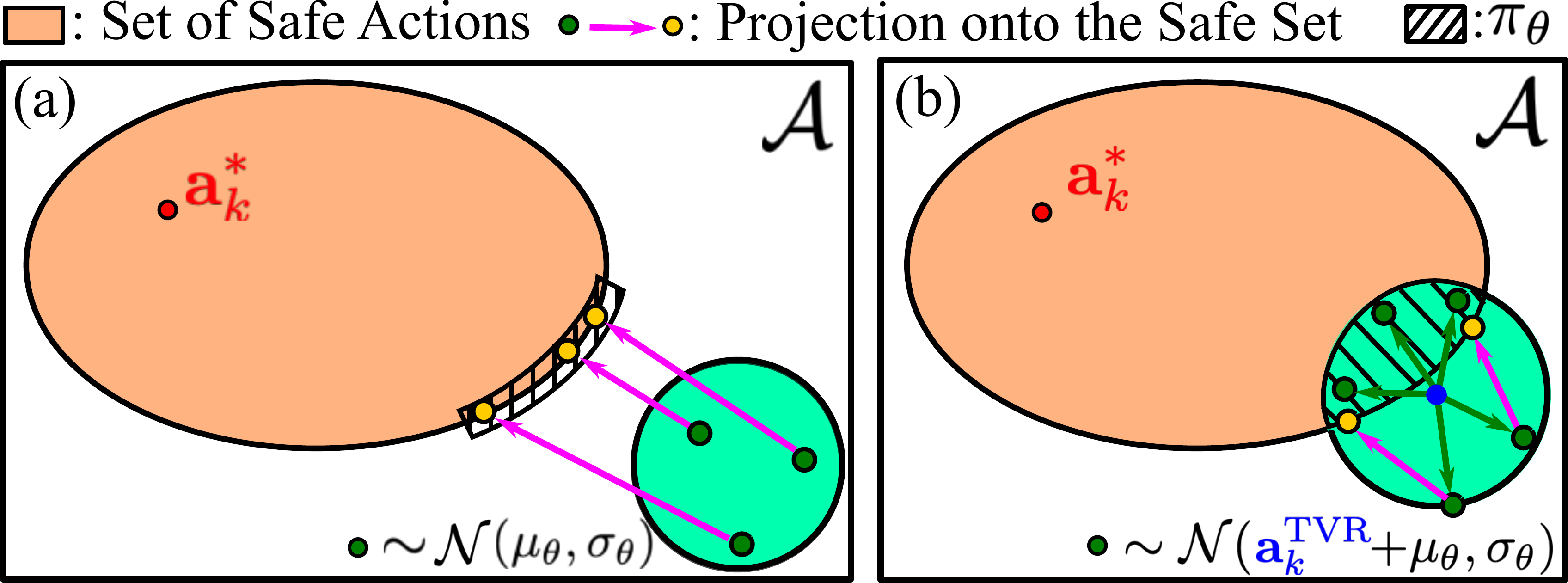}
    \caption{The safety compensation process. $\mathbf{a}_k^{*}$ denotes an optimal control input and the orange area represents a set of safe actions that ensures that the state at the next time step stays inside the safe set $\mathcal{C}$. (a) and (b) represent two different instances of feedforward exploration.}
    \label{fig:action_visualize}
\end{figure}

Particularly, $\mathbf{a}^{\rm{TVR}}_{k}$ in the feedforward exploration provides learning guidance and resolves two major issues in the safety projection: (1) inactive exploration and (2) the credit assignment problem. Consider, for example, two cases with different feedforward explorations, as illustrated in Fig.~\ref{fig:action_visualize}, whose final control policies are: (a) $\mathbf{a}_k = \mathbf{a}^{\bm{\theta}}_k + \mathbf{a}^{\rm{SF}}_k(\mathbf{a}^{\bm{\theta}}_k)$ and (b) $\mathbf{a}_k = \mathbf{a}^{\rm{TVR}}_k + \mathbf{a}^{\bm{\theta}}_k + \mathbf{a}^{\rm{SF}}_k(\mathbf{a}^{\rm{TVR}}_k + \mathbf{a}^{\bm{\theta}}_k)$.

In the case of (a) (and (b), respectively), the cyan area represents feedforward exploration expressed by a Gaussian distribution $\mathcal{N}(\bm{\mu}_{\bm{\theta}},\bm{\sigma}_{\bm{\theta}})$ (and $\mathcal{N}(\mathbf{a}_k^{\rm{TVR}}+\bm{\mu}_{\bm{\theta}},\bm{\sigma}_{\bm{\theta}})$, respectively), and the green dots are its samples. The pink arrow represents the safety compensation $\mathbf{a}_{k}^{\rm{SF}}(\mathbf{a}_{k}^{\bm{\theta}})$ (and $\mathbf{a}_{k}^{\rm{SF}}(\mathbf{a}_k^{\rm{TVR}}+\mathbf{a}_{k}^{\bm{\theta}}$), respectively). The black striped area is a distribution of the final action $\mathbf{a}_{k}$, and the yellow dots are its sample.

As Fig.~\ref{fig:action_visualize}(a) shows, there is no intersection between the set of safe actions and the possible feedforward exploration and the feedforward explorations are all projected onto the safe action set. The projection does not preserve the volume in the action space, and it hinders active explorations in the learning. However, Fig.~\ref{fig:action_visualize}(b) leverages the TVR planner as learning guidance and retains the volume in action space to explore over. When it comes to computing a gradient of the long-term reward, the projected actions make it difficult to evaluate the resulting trajectories and assign the credits in the $\bm{\theta}$ space. In other words, as Fig.~\ref{fig:action_visualize}(a) shows, three compensated samples (yellow dots) do not roll out different trajectories, which prevents the gradient descent and results in a local optimum.

\begin{table*}[bp]
\caption{Simulation Parameters}
\label{tab:parameters}
\begin{center}
\begin{tabular}{|c||c|c|c|c|c|c|c|c|c|c|c|c|c|c|c|c|}
\hline
 & \multicolumn{2}{c|}{LIPM} & \multicolumn{2}{c|}{$\mathcal{SM}$} & \multicolumn{2}{c|}{$\mathbf{a}^{\rm{TVR}}$} & $\mathbf{a}^{\bm{\theta}}$ & \multicolumn{2}{c|}{$\mathbf{a}^{\rm{SF}}$} & \multicolumn{5}{c|}{Reward} & \multicolumn{2}{c|}{Behavior} \\ \cline{2-17} 
\multirow{-2}{*}{} & \multicolumn{2}{c|}{$[h,l_{\rm{max}}]$} & \multicolumn{2}{c|}{$[T_{LN},T_{LF}]$} & \multicolumn{2}{c|}{$[T_{x'},T_{y'},\kappa_{x},\kappa_{y}]$} & Layer & \multicolumn{2}{c|}{$[K_\epsilon,\eta]$} & $r_a$ & $w_{b}$ & $w_{t}$ & $w_{s}$ & $w_{c}$ & \multicolumn{2}{c|}{$[\dot{x}^d,\omega_z^d]$} \\ \Xhline{2\arrayrulewidth}
\begin{tabular}[c]{@{}c@{}}DRACO\\ Walking \end{tabular} & \multicolumn{2}{c|}{$[0.93,0.7]$} & \multicolumn{2}{c|}{$[0.16,0.16]$} & \multicolumn{2}{c|}{$[0.22, 0.22,-0.18,-0.18]$} & $[64, 64]$ & \multicolumn{2}{c|}{$[10^5,0.8]$} & $5.0$ & $3.0$ & $3.0$ & $1.0$ & $1.0$ & \multicolumn{2}{c|}{$[0.3,0]$} \\ \hline
\begin{tabular}[c]{@{}c@{}}ATLAS\\ Walking\end{tabular} & \multicolumn{2}{c|}{$[0.82,0.55]$} & \multicolumn{2}{c|}{$[0.23,0.23]$} & \multicolumn{2}{c|}{$[0.15, 0.15,-0.16,-0.16]$} & $[64, 64]$ & \multicolumn{2}{c|}{$[10^5,0.8]$} & $5.0$ & $3.0$ & $3.0$ & $1.0$ & $1.0$ & \multicolumn{2}{c|}{$[0.15,0]$} \\ \hline
\begin{tabular}[c]{@{}c@{}}ATLAS\\ Turning\end{tabular} & \multicolumn{2}{c|}{$[0.82,0.55]$} & \multicolumn{2}{c|}{$[0.23,0.23]$} & \multicolumn{2}{c|}{$[0.15, 0.15,-0.16,-0.16]$} & $[64, 64]$ & \multicolumn{2}{c|}{$[10^5,0.8]$} & $5.0$ & $5.0$ & $5.0$ & $3.0$ & $1.0$ & \multicolumn{2}{c|}{$[0,0.09]$} \\ \hline
\end{tabular}
\end{center}
\end{table*}

\section{Simulation Results}
\label{sec:simulation_results}

We execute a series of experiments with our 10-DoF DRACO biped \cite{JunhyeokDRACO} and the 23-DoF Boston Dynamic's ATLAS humanoid using the DART simulator \cite{lee2018dart} to evaluate the proposed MDP formulation and policy design. The parameters used in the simulations are summarized in the Table~\ref{tab:parameters}. The goal of the experiments is three-fold: (1) How does the proposed method learn locomotion better than the baseline approaches (i.e., end-to-end policy search in \cite{schulman2017proximal}, DeepLoco in \cite{deeploco}, and the TVR planner) in terms of data-efficiency, safety, and the quality of the walking behavior? (2) How does each policy component in Eq.~\eqref{eq:action_design} contribute to the learning process? (3) Could the proposed approach be generalized to various types of walking (e.g., turning, walking over irregular terrain, and walking given random disturbances)?

\subsection{Forward Walking}
\subsubsection{Experiment Setup}
\begin{figure}[h]
    \centering
    \includegraphics[width=0.7\linewidth]{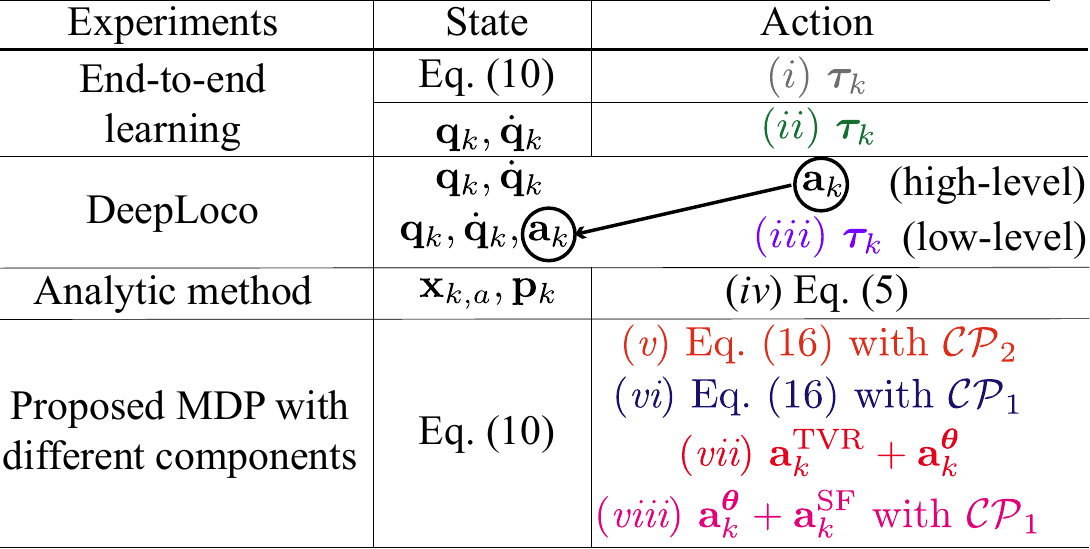}
    \caption{Eight MDPs with different states and actions for the forward walking.}
    \label{fig:experiment_setup}
\end{figure}

We include eight MDPs with different states and actions, and train policies for forward walking to demonstrate the effectiveness of our method. As baselines, two MDPs are based on an end-to-end model free learning: each policy learns joint torques $\bm{\tau}_{k}$ either from joint positions and velocities $\mathbf{q}_k,~\dot{\mathbf{q}}_{k}$ or from the LIPM described in Eq.~\eqref{eq:state}. We implement and adapt another baseline MDP from DeepLoco \cite{deeploco}, where a walking policy is composed of a high-level footstep planner and a low-level feedback controller. Note that DeepLoco trains the networks to generate footsteps and joint commands in an end-to-end manner, whereas our method incorporates a simplified model to train footstep policy efficiently. Another difference is that we consider a model-based feedback controller (i.e., WBC) to compute joint commands. The other baseline MDP is set up based on the deterministic policy shown Eq.~\eqref{eq:tvr} that maps LIPM information to footstep locations. Finally, we formulate four variations of our proposed MDP by alternating the components of the actions shown in Eq.~\eqref{eq:action_design}. Fig.~\ref{fig:experiment_setup} summarizes different states and actions for our experiments. We solve the MDPs using the policy search method described in \cite{schulman2017proximal} with the reward defined in Eq.~\eqref{eq:rew} except for the DeepLoco baseline where we follow the reward function and the actor-critic method described in \cite{deeploco}.
Experiments whose policy is a footstep location are followed by a low-level WBC. For example, a cubic spline trajectory is generated with a footstep decision made by the MDPs and converted to an operational space task. At the same time, a CoM position and torso orientation task are also specified with different priorities to maintain the robots upright.

\begin{figure}[t]
    \centering
    \includegraphics[width=1.0\linewidth]{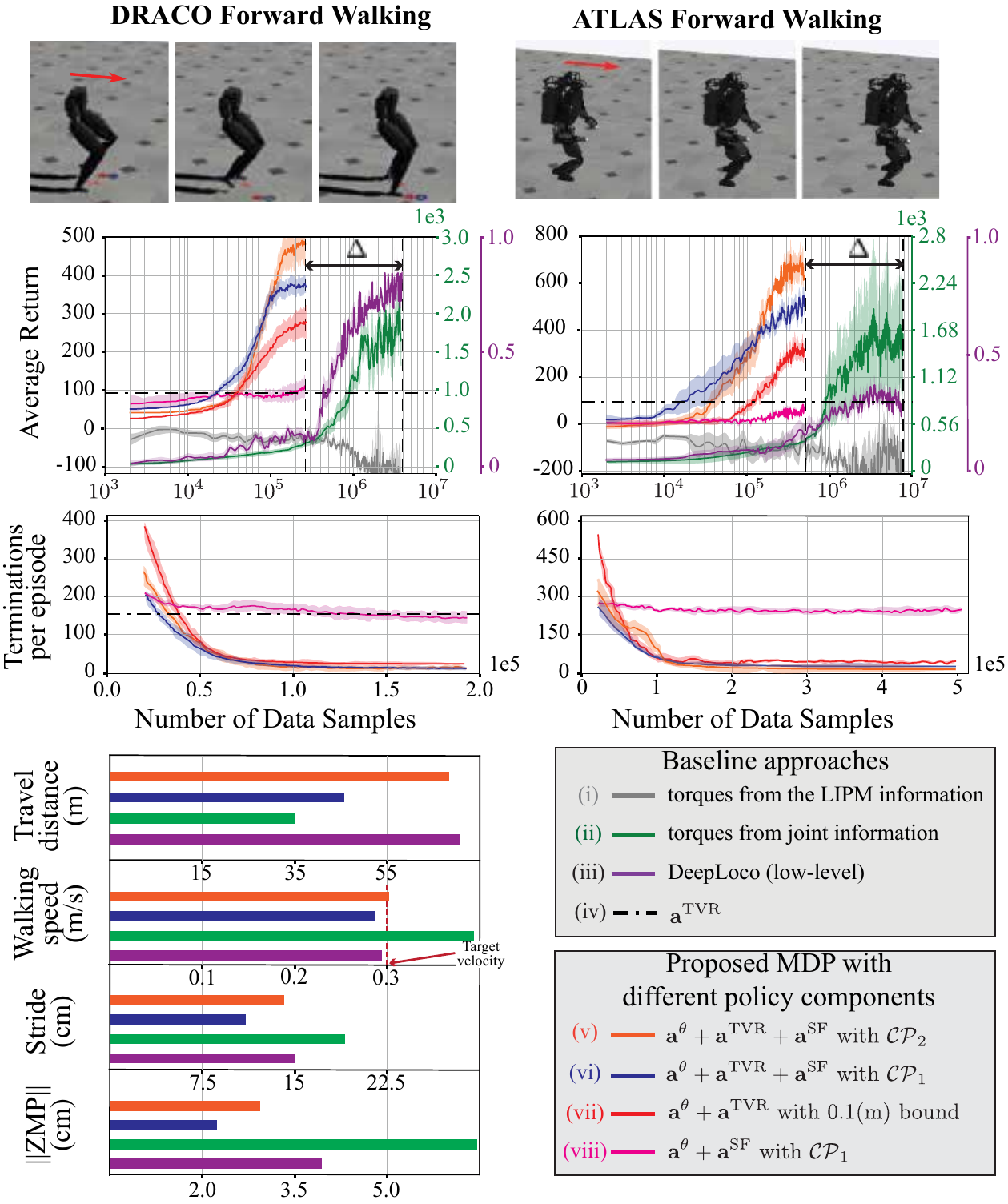}
    \caption{Learning curves for the experiments are shown here demonstrating learning performance for forward walking. The average return, the number of terminations per episode are shown throughout the training. Each of the curves is plotted with its mean and standard deviation across five runs. Note that in the average return plot, the green and gray curves use the green vertical axis, and the purple curve uses the purple vertical axis on the right side. At the bottom, we run an episode with trained policies from different setups and show the travel distance, walking speed, walking stride, and the average of the two-norm of the ZMP in the local frame.}
    \label{fig:learning_process}
\end{figure}

\subsubsection{Analysis}
Multiple policies are trained in each setup to regulate forward walking. The learning processes of the proposed MDP, as well as the baseline performance of the TVR and end-to-end learning, are illustrated in Fig.~\ref{fig:learning_process} with some useful metrics.

In the average return plot, the end-to-end learning with the LIPM information (gray curve) cannot achieve the motion of walking, whereas the other end-to-end learning with the joint information (green curve) shows a convergence of the walking behavior to unnatural motions. This shows that the LIPM information itself is not informative enough to calculate joint torques in an end-to-end manner. It is worth mentioning that the end-to-end learning with joint state information takes a more substantial dataset (denoted by $\Delta$) to generate desired locomotion behavior than using the proposed MDP. DeepLoco (purple curve) shows a faster convergence rate than the end-to-end learning in the case of DRACO thanks to the hierarchical policy structure. However, DeepLoco requires more data than our approach since it has to train the low-level feedback controller instead of using the WBC. Furthermore, DeepLoco does not scale well to ATLAS in our implementation.

The proposed MDP using the conservative one-step capture region (blue curve) helps to accelerate the learning at the beginning phase, but the one using the relaxed two-step capture region (orange curve) eventually achieves a better walking policy in terms of the average return. Training with a heuristic bound ($0.1 \si{\metre}$) instead of using the safety projection (red curve) exhibits relatively good performance, whereas the one without the TVR planner (pink curve) rarely improves throughout the updates. The results reflect the issues addressed in Section~\ref{sec:further_details}. The number of terminations per episode decays as the uncertain parts of the dynamics are revealed throughout the training. 

We evaluate the quality of walking resulting from different setups. In Fig.~\ref{fig:learning_process}, the trained policy from the blue curve uses conservative safety criteria, which results in smaller strides and slower walking speed than for the other methods. The walking behavior resulting from the green curve policy takes longer strides with faster walking speeds. However, as we can infer from the ZMP graph, the policy from the green curve shows unnatural walking motions and yields a short travel distance per episode.

\subsection{Generalization to various types of locomotion}
\subsubsection{Experiment Setup} 

We consider three additional experiments in simulation to show that our proposed formulation can be generalized to various types of locomotion: turning, walking over irregular terrain, and walking given random disturbances. For the turning experiment, the low-level WBC controls the robot's torso, pelvis, and feet orientation. We consider irregular terrains including tilted ground at angles of between $-10 \si{\degree}$ and $10\si{\degree}$. In addition, random disturbances are applied at intervals of $0.1\si{\sec}$ in the lateral and sagittal directions with a magnitudes between $-600 \si{\newton}$ and $600\si{\newton}$. Note that we apply the disturbances both before and after the \textit{Apex Moment}. For all experiments, the states, actions, and reward function are identical to the MDP formulation we described in Section \ref{sec:MDP_Formulation}. We train the policies using the policy search method described in \cite{schulman2017proximal}.

\begin{figure}[t]
    \centering
    \includegraphics[width=1.0\linewidth]{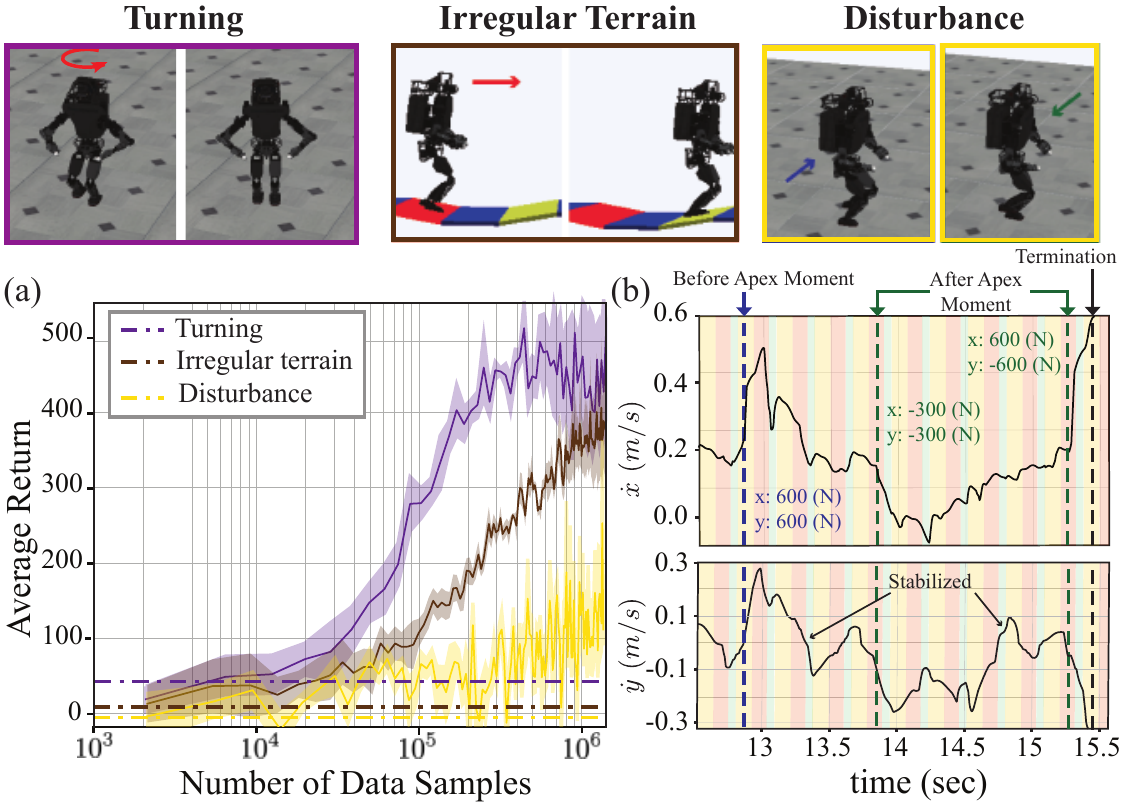}
    \caption{Various types of locomotion behaviors: turning, walking over irregular terrains, and walking with random disturbances. (a) The average return is shown throughout the training. (b) The trained policy for the disturbance experiments is evaluated. The CoM velocities are plotted when three different disturbances occur. Note that the background colors represent the locomotion states in Fig.~\ref{fig:preliminary}.}
    \label{fig:res2}
\end{figure}

\subsubsection{Analysis}
The learning processes for each experiment, as well as the baseline performance of the TVR planner, are illustrated in Fig.~\ref{fig:res2}(a). Our proposed approach succeeds in achieving locomotion behaviors based on average returns. The stance foot orientation captures the heading of the robot and the terrain information, and the neural network used in the learning process adapts to walking in new environments.

The experiment of walking with random disturbances shows an increment of the average return with high variance. It demonstrates robust walking under mild disturbances, but the average return is not as high as the return without disturbances shown in Fig.~\ref{fig:learning_process}. For further analysis, we evaluate a trained policy in the presence of three different types of disturbances: (1) a disturbance with a magnitude of $600 \si{\newton}$ on both the sagittal and the lateral directions before the \textit{Apex Moment} (i.e., in $\mathfrak{L}_{LF}$), (2) a disturbance with a magnitude of $300 \si{\newton}$ on both the sagittal and the lateral directions after the \textit{Apex Moment} (i.e., in $\mathfrak{L}_{LN}$), and (3) a disturbance with a magnitude of $600 \si{\newton}$ on both the sagittal and the lateral directions after the \textit{Apex Moment} (i.e., in $\mathfrak{L}_{LN}$). The velocity profiles of the CoM are shown in Fig.~\ref{fig:res2}(b). The first type of disturbance is dealt with by the robot using a single footstep. In the figure, one can see the CoM velocity in the lateral direction ($\dot{y}$) being directed back to near zero in the next double support phase, even with the large disturbances.  The second type of disturbance cannot be rejected by using a single step because that footstep is determined at the \textit{Apex Moment} that precedes the disturbance. However, this can still be compensated for in future walking steps, unless the magnitude of the disturbance is significant enough to make the robot fall immediately. In the last case, the magnitude of the disturbance is $600 \si{\newton}$ and makes the robot fall right away. Future work includes incorporating a disturbance observer and continuous disturbance detection during the swing motion as described in \cite{captur_point}. In such a case, the policy described in Eq.~\eqref{eq:action_design} will re-calculate new footstep locations to reject all disturbance within the first footstep.
\section{Concluding Remarks}
\label{sec:concluding_remarks}

In this letter, we describe an MDP formulation for data-efficient and safe learning for locomotion. Our formulation combines analytic and data-driven approaches to make high-level footstep decisions based on the LIPM. The proposed policy includes a TVR planner, a neural network, and a safety controller. The TVR planner computes achievable sub-optimal guidance, the neural network modulates the guidance to maximize the long-term reward, and the safety controller facilitates safe exploration during the learning process. The safety controller learns the unknown part of dynamics in tandem with the policy updates and compensate for unsafe actions from the neural network based on the capturability metric and the use of control-barrier function. We thoroughly evaluate the effectiveness of the proposed method show how it could be generalized for various types of walking with two humanoids. Our contributions include: (1) a structured learning control method that mitigates the limited effect of using simple models and generates agile and robust locomotion, (2) a data-efficient and safe learning process to reinforce walking using a physics-based model, and (3) the scalability of the method to various types of humanoid robots and walking. In the near future, we plan to implement this framework into a real bipedal robot called DRACO. In the past, we have encountered many problems using the LIPM without a learning process causing complicated tuning procedures. We believe that the policy learning technique presented here will automatically determine the gap between the model and reality and will adjust the policy accordingly with minimal tuning.


\ifCLASSOPTIONcaptionsoff
  \newpage
\fi

\bibliographystyle{IEEEtran}
\bibliography{2020_RAL}

\end{document}